\documentclass[journal=jacsat,manuscript=article]{achemso}

\usepackage{chemformula} % Formula subscripts using \ch{}
\usepackage[T1]{fontenc} % Use modern font encodings
\usepackage{amsmath,amsfonts}
\usepackage{algorithmic}
\usepackage{algorithm}
\usepackage{array}
\usepackage[caption=false,font=normalsize,labelfont=sf,textfont=sf]{subfig}
\usepackage{textcomp}
\usepackage{stfloats}
\usepackage{url}
\usepackage{verbatim}
\usepackage{graphicx}
\usepackage{booktabs}
\usepackage{tabularx}
\usepackage{diagbox}
\usepackage{multirow}
\usepackage{threeparttable}
\usepackage{xcolor}
\usepackage{mciteplus}

\author{Enqiang Zhu}
\author{Xiang Li}
\affiliation{Institute of Computing Science and Technology, Guangzhou University, Guangzhou 510006, China}

\author{Chanjuan Liu}
\affiliation{School of Computer Science and Technology, Dalian University of Technology, Dalian 116024, China}
\email{chanjuanliu@dlut.edu.cn}

\author{Nikhil R. Pal}
\affiliation{Electronics and Communication Sciences Unit, Indian Statistical Institute, Calcutta, India}

\title[Boosting drug-disease association prediction for drug repositioning via dual-feature extraction and cross-dual-domain decoding]
  {Boosting drug-disease association prediction for drug repositioning via dual-feature extraction and cross-dual-domain decoding}

\abbreviations{IR,NMR,UV}

\begin{document}


\begin{abstract}
The extraction of biomedical data has significant academic and practical value in contemporary biomedical sciences. In recent years, drug repositioning, a cost-effective strategy for drug development by discovering new indications for approved drugs, has gained increasing attention. However, many existing drug repositioning methods focus on mining information from adjacent nodes in biomedical networks without considering the potential inter-relationships between the feature spaces of drugs and diseases. This can lead to inaccurate encoding, resulting in biased mined drug-disease association information. To address this limitation, we propose a new model called Dual-Feature Drug Repurposing Neural Network (DFDRNN). DFDRNN allows the mining of two features (similarity and association) from the drug-disease biomedical networks to encode drugs and diseases. A self-attention mechanism is utilized to extract neighbor feature information. It incorporates two dual-feature extraction modules: the single-domain dual-feature extraction (SDDFE) module for extracting features within a single domain (drugs or diseases) and the cross-domain dual-feature extraction (CDDFE) module for extracting features across domains. By utilizing these modules, we ensure more appropriate encoding of drugs and diseases. A cross-dual-domain decoder is also designed to predict drug-disease associations in both domains. Our proposed DFDRNN model outperforms six state-of-the-art methods on four benchmark datasets, achieving an average AUROC of 0.946 and an average AUPR of 0.597. Case studies on two diseases show that the proposed DFDRNN model can be applied in real-world scenarios, demonstrating its significant potential in drug repositioning. 
\end{abstract}

\section{Introduction}
Drug development is a significant yet complex and time-consuming process. It consists of three main stages: the discovery stage, the preclinical stage, and the clinical stage. Typically, the entire process takes about 10 to 15 years and costs around \$2.8 billion on average \cite{fisher2006alterations,vijayan2022enhancing}. Despite ongoing efforts and innovations, the traditional drug development model faces considerable risks, often resulting in project terminations due to concerns about drug efficacy, safety, or commercial viability. Notably, a high percentage (80-90\%) of these projects fail during clinical trials. Consequently, finding a more efficient and precise method for drug development is critically important and valuable.

Given the high cost of drug development, computational drug repositioning has become a powerful strategy for discovering alternative uses of already approved drugs while considering their potential effects \cite{jarada2020review}. This approach utilizes computational and bioinformatics tools, providing greater efficiency and cost-effectiveness compared to traditional drug research and development methods. It enables the rapid screening of numerous candidate drugs and accelerates the development of new therapeutic options. Repurposing existing drugs can significantly lower development costs when compared to creating new drugs from scratch. In response to health crises, such as pandemics, repurposing known drugs offers a quicker solution than developing entirely new treatments. Furthermore, this method can unveil new applications for drugs that may be effective against diseases that currently lack effective therapies, thereby addressing major health challenges.

Machine learning has been commonly used in computational drug repositioning by information extraction from biomedical data. Techniques like random forest (RF), support vector machine (SVM), and decision trees are often applied for this purpose \cite{cai2023machine}. For instance, \cite{screening} introduced a novel drug-target interaction screening framework called PUDTI, which utilized SVM to identify potential drug repositioning targets. \cite{shi2019lasso} proposed a method for predicting drug-target interactions using Lasso dimensionality reduction and a random forest classifier. \cite{yang2023mgcnrf} employed random forests in their final prediction step, while other techniques such as decision trees \cite{gradient}, matrix factorization\cite{yang2014drug,sadeghi2022NMF-DR,liu2024tiwmflp}, and logistic regression \cite{yu2020gtb} have also been developed.

In recent years, deep learning technology has become a powerful tool in bioinformatics, as evidenced by \cite{lv2023identification, wang2024rpi, ma2022kg, yang2023area}. Unlike traditional machine learning methods, deep learning techniques leverage deep architectures to effectively capture high-level latent representations without requiring manual feature selection and tuning \cite{yu2022deep}. Existing deep learning methods for drug repositioning often incorporate graph neural networks like Graph Convolutional Networks (GCN) and Graph Attention Networks (GAT) to for model building \cite{lagcn,cai2021DRHGCN,sun2022PSGCN,tang2023DRGBCN,zhang2024NCH-DDA}. For instance, \cite{lagcn} performed graph convolution operations on the drug-disease heterogeneous network to encode drugs and diseases but did not account for the differences in encoding from different network spaces. \cite{cai2021DRHGCN} distinguished intra-domain and inter-domain feature extraction on the drug-disease heterogeneous network for encoding drugs and diseases. \cite{zhang2024NCH-DDA} extracted features of drugs and diseases from multiple neighborhood spaces and designed a contrastive learning loss to fuse these features. However, these studies did not consider calculating information weights from different neighbors. Authors in \cite{yang2024gcngat,huang2024HDGAT,du2024knowledge} used graph attention networks (GAT) to calculate the weights of neighbor aggregation and extract key neighbor information for encoding. Most deep learning methods have primarily focused on extracting valuable features from neighborhoods. However, these approaches often rely on a single encoding method, which does not intentionally account for relation prediction. As a result, a significant amount of noise can arise during model training, hindering the effective extraction of drug-disease association information and ultimately impacting the accuracy of the prediction results.

To address this challenge, we propose a new model called Dual-Feature Drug Repositioning Neural Network (DFDRNN). DFDRNN simultaneously uses two features, similarity and association, to encode drugs and diseases accurately. Compared to existing methods, the proposed model incorporates two dual-feature extraction modules: the SDDFE module for extracting features within a single domain (drug or disease) and the CDDFE module for extracting features across domains. DFDRNN uses a self-attention mechanism (SAM) to dynamically adjust the attention level of each input element to neighbor information, providing fine-grained weight allocation. This allows for efficient aggregation of neighbors and enhances the model's predictive power. A cross-dual-domain decoder is also designed to predict drug-disease associations in both domains. The key contributions of this work  are:

(1) To predict potential associations between drugs and diseases, we propose a dual-feature drug repositioning neural network (DFDRNN) model. A self-attention mechanism is used to capture complex relationships among adjacent nodes. The encoder of DFDRNN consists of two dual-feature extraction modules: the single-domain dual-feature extraction (SDDFE) module and the cross-domain dual-feature extraction (CDDFE) module. 

(2) The SDDFE module and the CDDFE module extract information from similarity networks and association networks, respectively, and track the dynamic changes of dual-feature to achieve precise encoding of drugs and diseases.

(3) A cross-dual-domain decoder is proposed to perform cross-computation between two encoded features. This process yields decoding results for both the drug and disease domains, which help effectively predict the drug-disease associations.

(4) The performance of the proposed method is compared with that of six state-of-the-art methods on four benchmark datasets. The experimental results show that DFDRNN outperforms other methods. In addition, case studies on Alzheimer's and Parkinson's diseases validated the reliability of DFDRNN in practical applications. 

The rest of this paper is structured as follows: Section \ref{section2} introduces the datasets and methods, providing details on the DFDRNN implementation; Section \ref{section3} explains the experimental setup and discusses the results; Section \ref{section4} concludes the study and suggests future research directions.

\section{Materials and Methods}\label{section2}
\subsection{Datasets}
We performed experiments on four benchmark datasets. The first, Gdataset \cite{Gdataset}, comprises 593 drugs from DrugBank \cite{knox2024drugbank} and 313 diseases from the Online Mendelian Inheritance in Man (OMIM) \cite{hamosh2005OMIM}, with a total of 1933 drug-disease associations. The second dataset, Cdataset \cite{BIRW_Cdataset}, consists of 663 drugs listed in DrugBank, 409 diseases from OMIM, and 2532 known associations. Ldataset \cite{lagcn}, the third dataset, contains 18,416 associations between 269 drugs and 598 diseases, derived from the Comparative Toxicogenomics Database (CTD) \cite{davis2023CTD}. Lastly, LRSSL \cite{liang2017lrssl} contains 763 drugs, 681 diseases, and 3051 validated drug-disease associations. A summary of the statistics for these datasets can be found in Table \ref{tab1}.

This study determines drug similarity based on chemical structures obtained from DrugBank using standardized SMILES notation \cite{weininger1988smiles}. For disease similarity, it is determined based on disease phenotypes and computed via MimMiner \cite{van2006text}, which measures similarity by analyzing the MeSH terms in the medical descriptions of diseases found in the OMIM database \cite{lipscomb2000medical}.

\begin{table}[h]
\caption{Summary statistics of the four datasets\label{tab1}}%
\centering
\begin{tabular*}{\columnwidth}{@{\extracolsep\fill}lccc@{\extracolsep\fill}}
\toprule
Datasets & No. of drugs  & No. of disease & No. of assocations\\
\midrule
Gdataset  & 593 & 313 & 1933  \\
Cdataset  & 663 & 409 & 2532  \\
Ldataset  & 269 & 598 & 18416 \\
LRSSL     & 763 & 681 & 3051  \\
\bottomrule
\end{tabular*}

\end{table}

\subsection{Model structure}
\begin{figure*}[!t]%
\centering
\includegraphics[width=15cm]{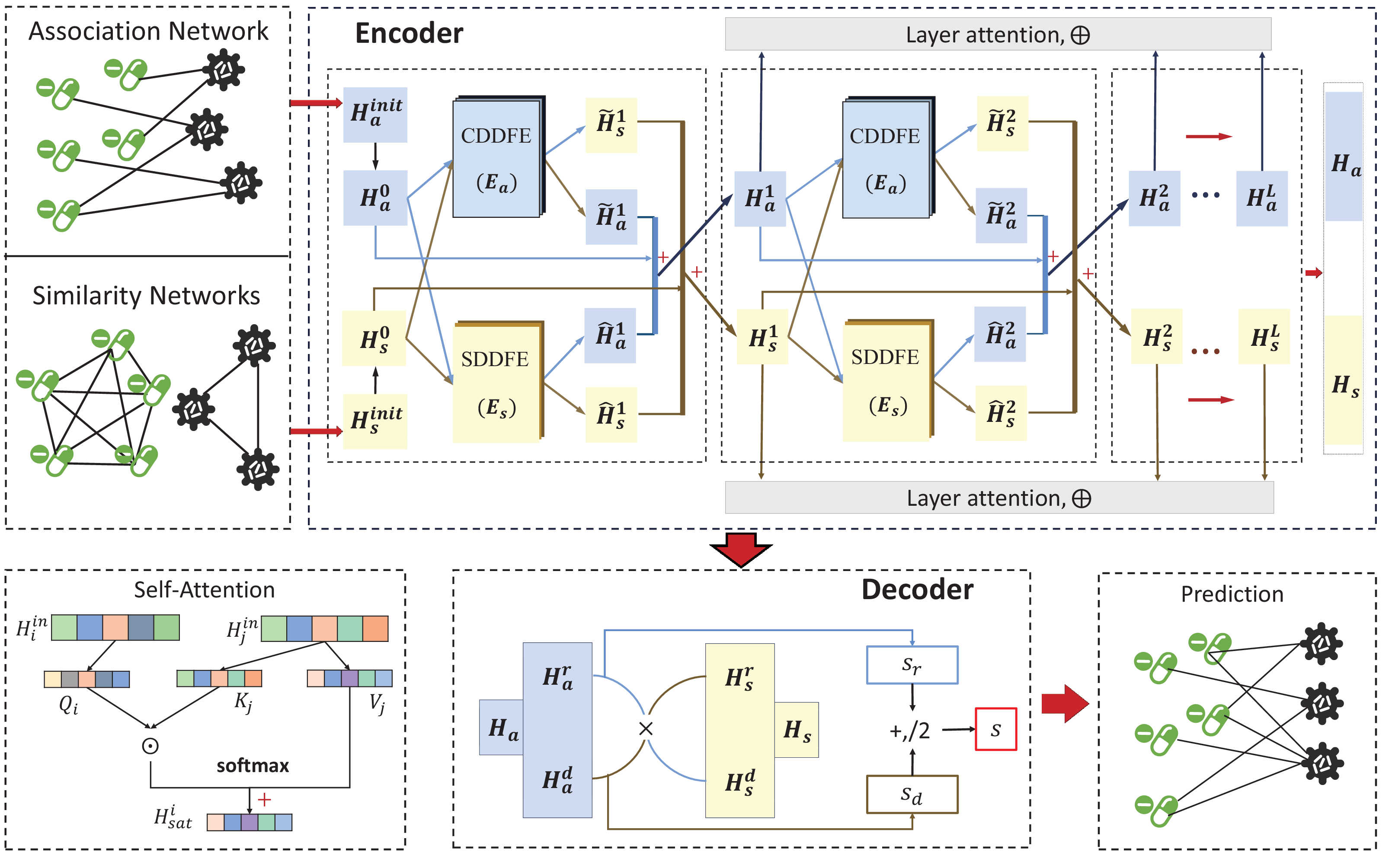}

\caption{DFDRNN generates the similarity feature and the association feature for drugs and diseases. The encoding process employs a multi-layer connected network architecture, where each layer performs SDDFE and CDDFE, while tracking the dynamic changes of dual-feature. A layer attention mechanism computes final features. The decoder then calculates drug-disease association scores separately for each domain.}\label{DFDRNN}
\end{figure*}

Our research is focused on predicting potential associations between drugs and diseases. To this end, we proposed a dual-feature drug repositioning neural network (DFDRNN) model. The encoder of DFDRNN encompasses two dual-feature extraction modules: the single domain dual-feature extraction (SDDFE) module and the cross domain dual-feature extraction (CDDFE) module. 
While the SDDFE module extracts features among drugs (or diseases), the CDDFE module extracts features between drugs and diseases and changes the features. A self-attention mechanism (SAM) model is employed when aggregating information to capture the complex relationship among adjacent nodes. The decoder of DFDRNN receives the results from the encoder and implements cross-dual-domain decoding, including both the drug and disease domains. Figure \ref{DFDRNN} illustrates the diagram of the proposed DFDRNN model.

\subsubsection{Three key networks}

To illustrate the relationships among biological entities (drugs and diseases) and describe their feature, we establish three networks: the drug-drug similarity network, a weighted complete graph with a node-set representing the drugs, and the weight of each edge representing the degree of similarity between the two drugs incident with the edge; the disease-disease similarity network, a weighted complete graph with node set representing the diseases and the weight of each edge representing the degree of similarity of the two diseases incident with the edge; and the drug-disease association network, a bipartite network with two parts representing the sets of drugs and diseases, respectively and each edge representing the associativity between the drug and diseases incident with the edge. We use three matrices to represent these three networks.  

Specifically, suppose that there are $n$ drugs  (denoted by $r_1,r_2,\ldots, r_n$) and $m$ diseases (denoted by $d_1,d_2,\ldots, d_m$). We use $S^{r}$ (or $S^{d}$) to denote the adjacency matrix of a drug-drug similarity network (or disease-disease similarity network), which is an $n\times n$ ($m\times m$) real matrix such that the entry $s^r_{ij}$ (or $s^d_{ij}$) $\in [0,1]$ denotes the degree of similarity of drug $i$ and drug $j$ (or disease $i$ and disease $j$), where $i,j\in \{1,2,\ldots, n\}$ (or $i,j\in \{1,2,\ldots, m\}$). For a drug-disease association network, we use a $n\times m$ binary matrix, denoted by $A$, to represent it, where the entry $a_{ij} =1$ if drug $i$ is associated with disease $j$ and $a_{ij} =0$ if drug $i$ is not associated with disease $j$ or the associativity is yet to be observed, for $i=1,2,\ldots, n$ and $j=1,2,\ldots, m$.

To reduce the impact of data noise, we aggregate only the top-$t$ neighbors for each drug in $S^{r}$ and each disease in $S^{d}$, and then two new matrices $R\in \mathbb{R}^{n\times n}$ and $D\in \mathbb{R}^{m\times m}$ are generated to replace $S^{r}$ and $S^{d}$, respectively, where the entries $r_{ij}$ ($i,j \in \{1,2,\ldots,n\}$) of $R$ and $d_{ij}$ ($i,j \in \{1,2,\ldots,m\}$) of $D$ are defined as follows. For a drug $r_i$ (and a disease $d_i$), we denote by $N_t(r_i)$ (and $N_t(d_i)$) the set of $t$-nearest neighbors of $r_i$ in $S^{r}$ (and $d_i$ in $S^{d}$), i.e., the weight of edge $r_ix$ (and edge $d_iy$) for every $x\in N_t(r_i)$ (and every $y\in N_t(d_i)$) is top-$t$ among all neighbors of $r_i$ (and $d_i$), and let  $\tilde{N_t}(r_i)=N_t(r_i)\cup \{r_i\}$ (and  $\tilde{N_t}(d_i)=N_t(d_i)\cup \{d_i\}$) be the extended $t$-nearest neighbors of $r_i$ (and $d_i$).

\begin{equation}\begin{matrix}  r_{ij} =\begin{cases} 
1,  & \mbox{if }r_j\in { \tilde{N_t}(r_i)} \\
0, & \text{otherwise}
\end{cases}\quad&
  d_{ij} =
\begin{cases} 
1,  & \mbox{if }d_j\in { \tilde{N_t}(d_i)} \\
0, & \text{otherwise}
\end{cases}
\end{matrix}
\end{equation}
\noindent 

Finally, to encode drugs and diseases simultaneously and distinguish SDDFE from CDDFE, we construct two heterogeneous networks $E_s$ and $E_a$ based on the above matrices $R, D$, and $A$. $E_s$ is used for SDDFE, while $E_a$ is used for CDDFE, defined as:

\begin{flalign}
&\ \qquad \qquad  E_s=\begin{bmatrix}
R     & 0      \\
0   & D
\end{bmatrix}\qquad   E_a=\begin{bmatrix}
0     & A      \\
A^T   & 0
\end{bmatrix}&
\end{flalign}
\subsubsection{Self-attention mechanism}
The self-attention mechanism (SAM) can be described as a process associated with three vectors, i.e., one query vector $Q_i$, some key vectors $K_j$, and some value vector $V_j$, which maps $Q_i$ and a group of $K_j$-$V_j$ pairs to a new output vector. The output of the SAM is a weighted sum of $V_j$ vectors, where the weight for each $V_j$ is computed by taking the dot product of the $Q_i$ with the corresponding $K_j$ \cite{vaswani2017attention}. In our SDDFE and CDDFE modules, we use the multi-head SAM to aggregate neighbor information for drugs and diseases. We use the notation $\mathbf{sam}$ to represent the multi-head SAM mapping that involves two parameters $E$ and $H^{in}$, where $E\in \{E_s, E_a\}$ and $H^{in} \in \mathbb{R}^{(n+m)\times k}$ is an input feature matrix with embedding dimension $k$. The specific implementation of $\mathbf{sam}$ can be described as follows: 

\begin{equation}  Q=H^{in}w^Q, \qquad K=H^{in}w^K, \qquad V=H^{in}w^V\end{equation}
where $w^Q, w^K, w^V \in \mathbb{R}^{\mathit k\times k}$ are three trainable $k\times k$ real matrices. Clearly, $Q, K,$ and $V$ are $(n+m)\times k$ real matrices. 

Each $Q$, $K$, and $V$ is partitioned into $p$ parts (called heads), denoted by $Q^i$,$K^i$, and $V^i$, $i=1,2,\ldots, p$, and the dimension of each part (i.e., the number of columns) is $k'=k/p$. We use $Q^q_i$, $K^q_i$, and $V^q_i$ to denote the feature of node $i$ in $Q^q$, $K^q$, and $V^q$ (i.e., the row vector corresponding to $i$), respectively. 
Let $X^q$ be the $q$-th head of $X\in \{Q,K,V\}$. Then, the weight of aggregating neighbor information is calculated based on $Q^q$ and $K^q$. Specifically, for a node $i$ (drug or disease), the weight $\beta^q_{ij}$ of $i$ aggregating its neighbor $j$ is defined as

\begin{equation}  \beta^q_{ij}=\frac{ exp(\frac{Q^q_i\bigodot K^q_j}{\sqrt{k'}})}{\sum_{z \in N(i)} exp(\frac{Q^q_i\bigodot K^q_z)}{\sqrt{k'}})}\end{equation}

\noindent where $\bigodot$ represents the inner product of vectors, $N(i)$ is the set of neighbors of node $i$. We denote by $\mathbf{concat} (x_1,x_2,\ldots,x_p)$ the concatenation operation of vectors $x_1,x_2,\ldots,x_p$. Then, the result corresponding to node $i$ obtained by the SAM mapping (i.e., the $i$-th row vector in $\mathbf{sam}(E,H^{in})$), denoted by $H_{\mathbf{sam}}^i$, can be described as  
\begin{equation}  H_{\mathbf{sam}}^i = \mathbf{concat} (\sum_{j \in N(i)}\beta^1_{ij}V^1_j,\sum_{j \in N(i)}\beta^2_{ij}V^2_j,\ldots,\sum_{j \in N(i)}\beta^p_{ij}V^p_j) 
\end{equation}

Moreover, let $H_{\mathbf{sam}}$ represent the feature matrix obtained by SAM, $H_r \in \mathbb{R}^{n\times k}$ be the drug feature obtained by SAM, and $H_d \in \mathbb{R}^{m\times k}$ is the disease feature obtained by SAM. Then, we have 

\begin{equation}
    H_r = \begin{bmatrix}H^{r_1}_{\mathbf{sam}}\\H^{r_2}_{\mathbf{sam}}\\ \dots \\H^{r_n}_{\mathbf{sam}}
    \end{bmatrix}, \hspace{0.3cm}
      H_d = \begin{bmatrix}H^{d_1}_{\mathbf{sam}}\\H^{d_2}_{\mathbf{sam}}\\ \dots  \\H^{d_m}_{\mathbf{sam}}
    \end{bmatrix}, \hspace{0.3cm}
    H_{\mathbf{sam}}=\begin{bmatrix}H_r\\H_d\end{bmatrix}  
\end{equation}

After the SAM function, a fully connected layer is employed to further extract the features of drugs and diseases. We use the notation $\mathbf{samf}$ to denote the whole process, i.e., $\mathbf{samf}$ is the combination of $\mathbf{sam}$ with a fully connected (FC) layer. Let $w_r\in \mathbb{R}^{k\times k}$ and $w_d\in \mathbb{R}^{k\times k}$ be trainable parameter matrices, $b_r$ and $b_d$ be biases, and $\sigma$ be a LeakyRELU \cite{maas2013leakyrelu} activation function. Then, 

{\begin{equation}
\begin{aligned}
     \mathbf{samf}(E, H^{in})& = \mathbf{FC}(\mathbf{sam}(E,H^{in}))\\
     &=\begin{bmatrix}\sigma(H_rw_r + b_r)\\\sigma(H_dw_d + b_d)\end{bmatrix}
\end{aligned}
\end{equation}}

\subsubsection{Encoder}
The DFDRNN model is a multi-layer connected network architecture, where each layer involves two types of dual-feature extraction modules: SDDFE module based on network $E_s$ and CDDFE module based on network $E_a$. 
First, the initial features of drugs and diseases are generated in two features, including the similarity feature $H^{init}_s$ and the association feature $H^{init}_a$, described as follows.
\begin{flalign}
&\   \qquad \qquad H^{init}_s=\begin{bmatrix}
S^r     & 0      \\
0   & S^d
\end{bmatrix}\qquad   H^{init}_a=\begin{bmatrix}
0     & A      \\
A^T   & 0
\end{bmatrix}&
\end{flalign}

Next, DFDRNN projects the initial features to a $k$-dimensional space through a linear layer. The similarity feature $H_s^0$ and the association feature $H_a^0$ in the $0$-layer are defined as follows:  

\begin{equation}\label{eq.project}
      H^0_s = H^{init}_s \times M \quad, \quad H^0_a = H^{init}_a \times M
\end{equation}
where $M\in \mathbb{R}^{(n+m)\times k}$ is trainable parameter matrix.

Now, we consider the process of extracting features in the $\ell$-th layer for $\ell\geq 1$. In the $\ell$-th layer, we denote by $f^{\ell}$ the SDDFE function in this layer, $H^{\ell}_s$ and $H^{\ell}_a$ are the input similarity feature and the input association feature, respectively,  $\hat{H}^{\ell+1}_{s}$ and $\hat{H}^{\ell+1}_{a}$ represent the obtained similarity feature and association feature by SDDFE, and $\mathbf{samf}^{\ell}_f$  the aggregation function of $\ell$-th layer for SDDFE. Then, 

\begin{equation}\label{eq.sddfe}
    \begin{aligned}
      (\hat{H}^{\ell+1}_{s}, \hat{H}^{\ell+1}_{a}) & =  f^\ell(E_s, H^{\ell}_s, H^{\ell}_a) \\
    &  = (\mathbf{samf}^{\ell}_f(E_s, H^{\ell}_s),\mathbf{samf}^{\ell}_f(E_s, H^{\ell}_a) )
\end{aligned}
\end{equation}

Since SDDFE extracts features within a single domain, the dual-feature does not undergo dynamic changes. Thus, the input similarity features $H^{\ell}_s$ produce the output similarity features $\hat{H}^{\ell+1}_s$, and the input association features $H^{\ell}_a$ produce the output association features $\hat{H}^{\ell+1}_a$. Next, we use $g^{\ell}$ to denote the CDDFE function in the $\ell$-th layer. Similarly, $\tilde{H}^{\ell+1}_{s}$ and $\tilde{H}^{\ell+1}_{a}$ represent the obtained similarity feature and association feature by CDDFE, respectively, and $\mathbf{samf}^{\ell}_g$ represents the aggregation function.
Then,
\begin{equation}\label{eq.cddfe}
    \begin{aligned}
    (\tilde{H}^{\ell+1}_{a},\tilde{H}^{\ell+1}_{s}) & = {g^\ell(E_a, H^\ell_s, H^\ell_a)} \\
    &   {= (\mathbf{samf}^\ell_g(E_a, H^\ell_s),\mathbf{samf}^\ell_g(E_a, H^\ell_a))}
\end{aligned}
\end{equation}

Since the CDDFE module extracts features across domains, dual-feature undergo dynamic changes. Specifically, the similarity feature $H^\ell_s$ is changed into the association feature $\tilde{H}^{\ell+1}_a$, and the association feature $H^\ell_a$ is changed into the similarity feature $\tilde{H}^{\ell+1}_s$.

Finally, for each module, the similarity (and association) features  (including one input feature and two output features corresponding to the two modules) in the $\ell$-layer are added into an input similarity (and association) feature of the  $(\ell+1)$-layer, i.e.,
\begin{equation}\label{fuse.sddfe}
      H^{\ell+1}_s = \hat{H}^{\ell+1}_{s} + \tilde{H}^{\ell+1}_{s} + H^{\ell}_{s}
\end{equation}
\begin{equation}\label{fuse.cddfe}
      H^{\ell+1}_a = \hat{H}^{\ell+1}_{a} + \tilde{H}^{\ell+1}_{a} + H^{\ell}_{a}
\end{equation}

As the depth of network layers increases, the valuable information for predicting drug-disease association extracted by the DFDRNN model decreases. Consequently, a layer attention mechanism is incorporated to assign distinct weights to the outputs of each layer. We use $L$ to represent the total number of layers in DFDRNN. The weight of the $\ell$-th layer, denoted by $\beta^\ell$, is initialized by $1/L$ and automatically learned by the model. Then, the final encoding of the similarity feature and the association feature $H^{r}_s$, $H^{r}_a$, $H^{d}_s$, and $H^{d}_a$ of drugs and diseases are obtained by implementing the layer attention mechanism as follows.

\begin{equation}\label{final}
    \begin{bmatrix}
          H^{r}_s \\H^{d}_s
    \end{bmatrix} = \sum_{\ell=1}^L(\beta^\ell\times H^\ell_s), \qquad
    \begin{bmatrix}
          H^{r}_a \\H^{d}_a
    \end{bmatrix} = \sum_{\ell=1}^L(\beta^\ell\times H^\ell_a)
\end{equation}

\subsubsection{Decoder}
A cross-dual-domain decoder is utilized to predict drug-disease association. We perform cross computation between two features, using $s_r \in \mathbb{R}^{n\times m}$ and $s_d \in \mathbb{R}^{n\times m}$ to represent the prediction values for the drug domain and the disease domain, respectively. Let $s \in \mathbb{R}^{n \times m}$ denote the final score matrix of drug-disease associations, where $s_{ij}$ represents the prediction score between drug $r_i$ and disease $d_j$. That is
\begin{alignat}{2}
 {s_r}&= \mathbf{sigmod}({  H^{r}_a\times (H^{d}_s)^T})\label{r}\\
 {s_d}&=  \mathbf{sigmod}(  H^{d}_a\times (H^{r}_s)^T)\label{d}\\
 {s} &={  (s_r + (s_d)^T)/2.0}\label{s}
\end{alignat}
Algorithm $\ref{algo1}$ gives a detailed explanation of predicting novel drug-disease associations using DFDRNN.

\begin{algorithm*}[t]
\caption{DFDRNN}\label{algo1}
\begin{algorithmic}[1]
    \STATE \textbf{Input:} Two heterogeneous network ($E_s$ and $E_a$); initial the similarity feature and the association feature (${H}^{init}_{s}$ and ${H}^{init}_{a}$); number of layers $L$; maximum training epochs $C$.
    
    \STATE \textbf{Output:} Drug-disease association score matrix $s$.
    \STATE Initialize the weight matrix ($  w^Q,w^K,w^V,w_r,w_b,\ldots$).
    \FOR {$c = 1$ to $  C$}
        \STATE \hspace{0.5cm} Compute the similarity feature $H^0_s$ and the association feature $H^0_a$ with Equation \eqref{eq.project}.
        \STATE \hspace{0.5cm} \FOR {$\ell =  $ to $  L-1 $}
            \STATE \hspace{0.5cm} \hspace{0.5cm} Compute SDDFE $  \hat{H}^{\ell+1}_s,   \hat{H}^{\ell+1}_a$ with Equation \eqref{eq.sddfe}.
            \STATE \hspace{0.5cm} \hspace{0.5cm} Compute CDDFE $  \tilde{H}^{\ell+1}_s,  \tilde{H}^{\ell+1}_a$ with Equation \eqref{eq.cddfe}.
            \STATE \hspace{0.5cm} \hspace{0.5cm} Fusing the same feature with Equations \eqref{fuse.sddfe} and \eqref{fuse.cddfe} produces $H^{\ell+1}_s$ and $H^{\ell+1}_a$.
        \STATE \hspace{0.5cm} \ENDFOR
        \STATE \hspace{0.5cm} Final embedding via layer attention and Equation \eqref{final}.
        \STATE \hspace{0.5cm} Obtain Computer the prediction score matrix $s$ with Equation \eqref{r}\eqref{d}\eqref{s}
        \STATE \hspace{0.5cm} Update the parameters $ w^Q,w^K,w^V,w_r,w_b,\ldots$ with Equation \eqref{loss}.
        \STATE \hspace{0.5cm} $c \gets c + 1$.
    \ENDFOR
\end{algorithmic}
\end{algorithm*}

\subsection{Optimization}
Predicting drug-disease associations can be treated as a binary classification problem, where the goal is to determine if a specific drug is associated with a specific disease. Due to the significant imbalance between the number of positive samples (associations) and negative samples (non-associations), we utilize binary cross-entropy as the loss function. Let $y^+$ and $y^-$ represent the positive and negative labels in the dataset, respectively. To mitigate the effects of this imbalance, we introduce a weight $\lambda = \frac{|y^-|}{|y^+|}$, where $|y^{-}|$ and $|y^+|$ denote the sizes of the negative and positive sample sets. A drug-disease pair $(i, j)$ corresponds to $r_i$ (drug) and $d_j$ (disease) in either $y^+$ or $y^-$, respectively.

\begin{equation}\label{loss}
      loss = - \frac{1}{n+m}(\lambda \times \sum_{(i,j)\in y^+}\log(s_{ij}) + \sum_{(i,j)\in y^-}\log(1 - s_{ij}))
\end{equation}

The DFDRNN model is optimized using the Adam optimizer \cite{kingma2014adam}, with parameters initialized through Xavier initialization \cite{glorot2010understanding}. To enhance generalization, we randomly drop edges and apply regular dropout techniques \cite{srivastava2014dropout}.

\section{Experiments and Results}\label{section3}

\subsection{Baseline methods}
We compared DFDRNN against six state-of-the-art methods for predicting drug-disease associations across the above four datasets. These methods include:

$\bullet$ LAGCN \cite{lagcn}: This method utilizes multi-layer GCNs for neighbor information propagation on heterogeneous networks and introduces the layer attention mechanism.

$\bullet$ DRHGCN \cite{cai2021DRHGCN}: This model uses GCN and bilinear aggregators to design inter-domain and intra-domain embeddings for predicting drug-disease associations.

$\bullet$ DRWBNCF \cite{meng2022DRWBNCF}: This model focuses on the interaction information between neighbors and uses generalized matrix factorization collaborative filtering to predict drug-disease associations.

$\bullet$ NCH-DDA \cite{zhang2024NCH-DDA}: This method involves the design of a single-neighborhood feature extraction module and a multi-neighborhood feature extraction module to extract features for drugs and diseases. Additionally, it introduces neighborhood contrastive learning for drug-disease association prediction.

$\bullet$ HDGAT \cite{huang2024HDGAT}: This approach combines graph convolutional neural networks with bidirectional extended short-term memory networks and integrates layer attention mechanisms and residual connections.

$\bullet$ DRGBCN \cite{tang2023DRGBCN}: This method utilizes layer attention graph convolutional networks to encode drugs and diseases and employs bilinear attention network modules to enhance their intricate relationship.

\subsection{Experimental parameter configuration}
The model's parameter values are as follows: DFDRNN's feature embedding dimension is 128; the number of network layers is 3; the number of attention heads is p = 2; the dropout rate is 0.4; the edge dropout rate is 0.5 on the Ldataset and 0.2 on other datasets; the learning rate is 0.008; and the maximum training epoch is 800. Additionally, we set top-$t$ to 7 based on parameter sensitivity experiments.

\begin{table*}[!t]
\caption{AUROC and AUPR results for DFDRNN and comparative models using 10-times repeated 10-fold cross-validation \label{auroc}}
\tabcolsep=0pt%%
\centering
\begin{tabular*}{0.95\textwidth}{@{\extracolsep{\fill}}llccccl@{\extracolsep{\fill}}}
\toprule%
Metrics& Model & Gdataset & Cdataset & Ldataset & LRSSL & Avg.\\
\midrule
\multirow{7}*{AUROC}
& LAGCN   & 0.881±0.002 & 0.911±0.002 & 0.864±0.001 & 0.935±0.001 & 0.898\\
& DRHGCN  & 0.949±0.001 & \underline{0.965±0.001} & 0.864±0.001 & 0.960±0.001 & 0.935\\
& DRWBNCF & 0.924±0.002 & 0.942±0.001 & 0.821±0.001 & 0.935±0.001 & 0.906\\
& NCH-DDA & \underline{0.954±0.002} & 0.959±0.002 & \underline{0.879±0.001} & \underline{0.963±0.002} & \underline{0.939}\\
& HDGAT   & 0.917±0.002 & 0.939±0.001 & 0.874±0.001 & 0.936±0.001 & 0.916\\
& DRGBCN  & 0.931±0.003 & 0.946±0.001 & 0.824±0.002 & 0.943±0.001 & 0.911\\
& DFDRNN  & \textbf{0.960±0.001} & \textbf{0.973±0.001} & \textbf{0.885±0.001} & \textbf{0.965±0.001} & \textbf{0.946}\\
\midrule
\multirow{6}*{AUPR}
& LAGCN   & 0.197±0.005 & 0.307±0.008 & 0.522±0.002 & 0.254±0.002 & 0.320\\
& DRHGCN  & \underline{0.531±0.003} & 0.619±0.003 & 0.523±0.001 & 0.408±0.002 & 0.520\\
& DRWBNCF & 0.493±0.001 & 0.569±0.004 & 0.413±0.003 & 0.349±0.005 & 0.456\\
& NCH-DDA & \textbf{0.623±0.006} & \underline{0.675±0.003} & \underline{0.552±0.002} & \underline{0.466±0.002} & \underline{0.579}\\
& HDGAT   & 0.505±0.005 & 0.566±0.008 & 0.546±0.003 & 0.403±0.005 & 0.505\\
& DRGBCN  & 0.408±0.007 & 0.451±0.006 & 0.434±0.005 & 0.262±0.006 & 0.388\\
& DFDRNN  & \textbf{0.623±0.003} & \textbf{0.688±0.002} & \textbf{0.573±0.001} & \textbf{0.502±0.002} & \textbf{0.597}\\
\bottomrule

\end{tabular*}
\begin{tablenotes}%
\item The highest values in each column are highlighted in bold, while the second-highest values are underlined
\end{tablenotes}
\end{table*}

\subsection{Evaluation metrics}

We classify the prediction outcomes into four categories. As illustrated in Table \ref{Evaluate}, TP denotes true positives that are correctly identified as positive, FN represents false negatives that are wrongly classified as negative, FP indicates false positives that are incorrectly classified as positive, and TN stands for true negatives that are accurately identified as negative. Based on these definitions, we proceed to calculate the True Positive Rate (TPR or Recall), False Positive Rate (FPR), and Precision using the following formulas:

\begin{table}[!h]
    \caption{The statistics of the four categories\label{Evaluate}}%
    \centering
    \begin{tabular*}{0.9\columnwidth}{@{\extracolsep{\fill}}c|c|c}
    \hline
    \diagbox{Actual}{Predicted} & Positive\hspace{1cm} & Negative\hspace{1cm} \\
    \hline
    Positive & TP\hspace{1cm} & FN\hspace{1cm} \\
    \hline
    Negative & FP\hspace{1cm} & TN\hspace{1cm} \\
    \hline
    \end{tabular*}
\end{table}

\begin{equation}
    \text{TPR(or Recall)} = \frac{TP}{TP+FN}
\end{equation}
\begin{equation}
    \text{FPR} = \frac{FP}{FP+TN}
\end{equation}
\begin{equation}
    \text{Precision} = \frac{TP}{TP+FP}
\end{equation}

By calculating the FPR, TPR, and precision across different thresholds, we generate the receiver operating characteristic (ROC) curve, plotting FPR on the X-axis and TPR on the Y-axis. Similarly, the precision-recall (PR) curve is created, with recall on the X-axis and precision on the Y-axis. The evaluation metrics for DFDRNN include the area under the ROC curve (AUROC) and the area under the PR curve (AUPR).

\subsection{DFDRNN's performance in cross-validation}

We utilize 10-fold cross-validation (10-CV) to evaluate the model's performance. The dataset is randomly divided into ten distinct subsets, each comprising 10\% known drug-disease associations and 10\% unknown pairs. In each iteration, one subset is used as the test set, while the other nine serve as the training set. This process is repeated ten times with varying random seeds, allowing for the calculation of the average and standard deviation of the performance metrics. The experimental outcomes, including AUROC and AUPR scores, are summarized in Table \ref{auroc}, while the corresponding ROC and PR curves are provided in Supplementary Figures S1-S4.

In Table \ref{auroc}, it is evident that DFDRNN outperformed all other models by achieving the highest average AUROC of 0.946 and the highest average AUPR of 0.597 across the four datasets. This performance surpassed the second-ranked NCH-DDA model by 0.7\% and 1.8\% in AUROC and AUPR, respectively. DFDRNN demonstrated the best AUROC and AUPR on all four datasets, showcasing its predictive solid capability for potential drug-disease associations based on known associations. It's worth mentioning that NCH-DDA performed exceptionally well, coming in second only to DFDRNN, which may be attributed to its feature extraction from different neighborhoods and the fusion of features using contrastive learning loss.

\subsection{Effect of the number of nearest neighbors}

\begin{figure*}[t]
\begin{center}
\subfloat[]{
\includegraphics[width=0.36\textwidth]{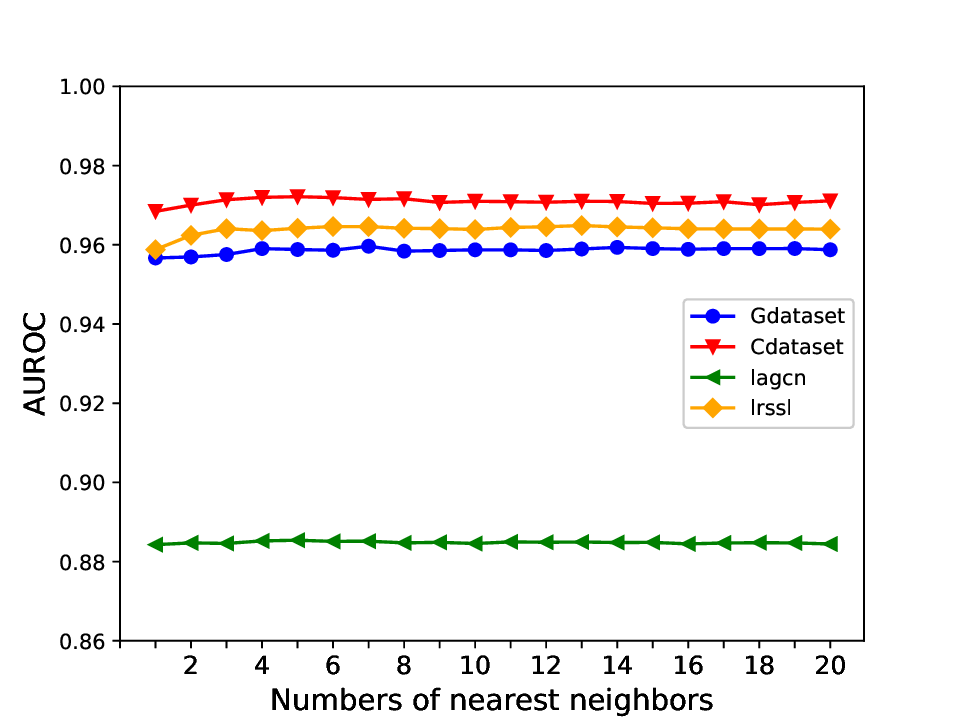}%
}
\subfloat[]{
\includegraphics[width=0.36\textwidth]{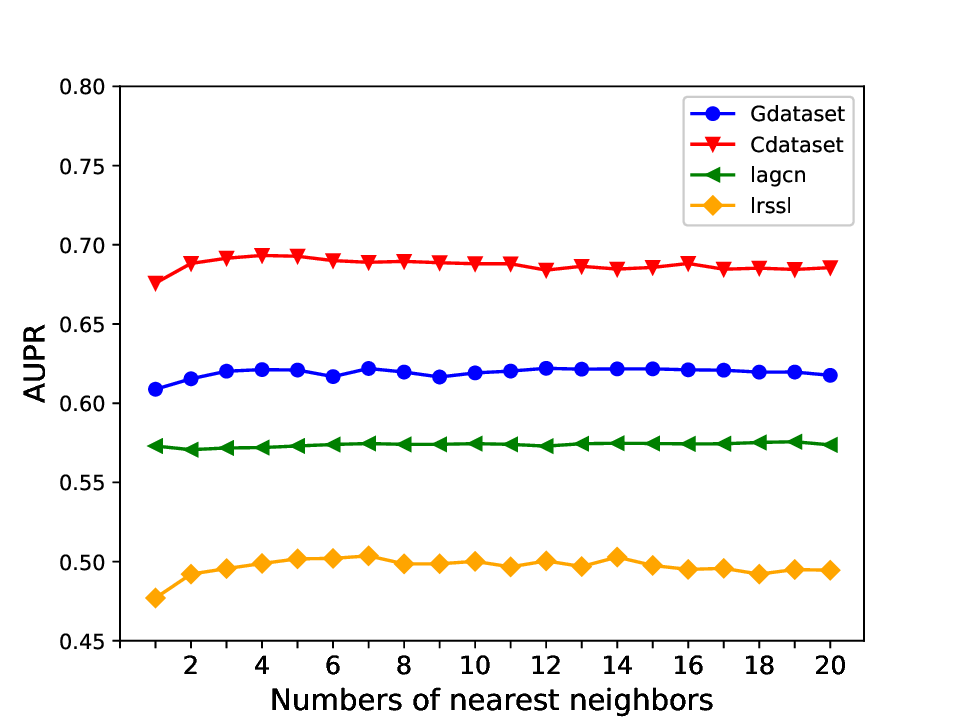}%
}
\hfil
\subfloat[]{
\includegraphics[width=0.36\textwidth]{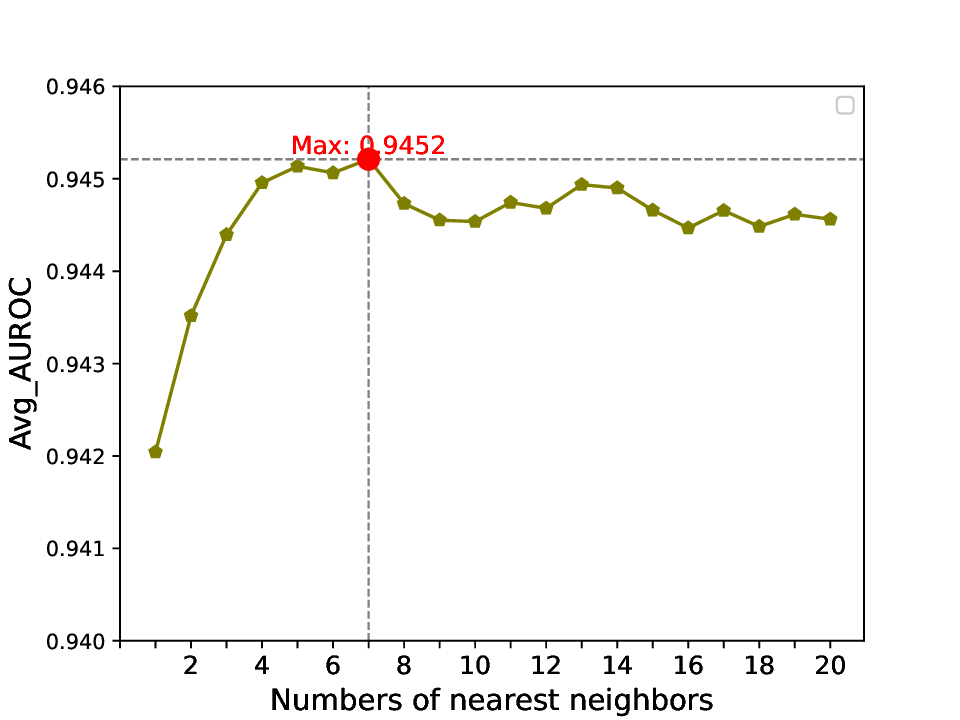}%
}
\subfloat[]{
\includegraphics[width=0.36\textwidth]{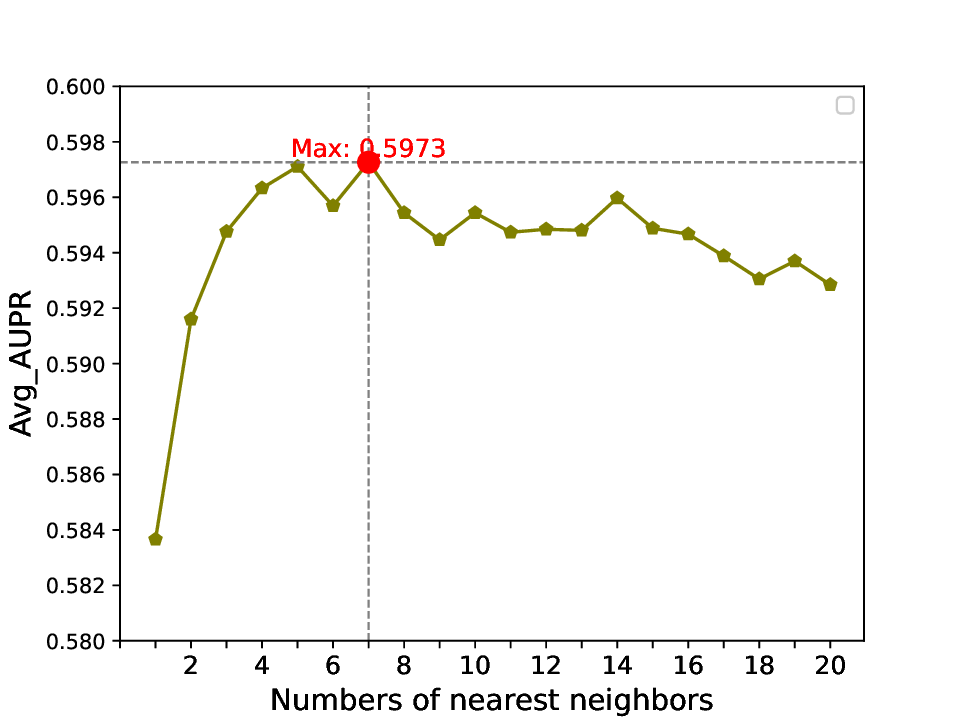}%
}
\caption{(a)Line plot of AUROC as a function of top-$t$ on four benchmark datasets; (b)Line plot of AUPR as a function of top-$t$ on four benchmark datasets; (c)Line plot of the average AUROC as a function of top-$t$ on four datasets; (d)Line plot of the average AUPR as a function of top-$t$ on four datasets.}\label{numk}
\end{center}
\end{figure*}

In the context of DFDRNN, the value of top-$t$ determines the number of neighbors features aggregated during SDDFE, which influences the model's performance. A small value of $t$ limits the neighbor information obtained, thereby affecting the model's performance. On the other hand, as $t$ increases, more neighbors are obtained, but this also introduces noise information, leading to the propagation of unnecessary data. The performance of DFDRNN under 10-CV varies with $t$ on the four datasets, as demonstrated in Figure \ref{numk} for $t\in \{1, 2, 3, \ldots, 20\}$. 

Based on the results shown in Figure \ref{numk} (a) and (b), DFDRNN performs the poorest across the four datasets when $t = 1$. However, as the $t$ value increases, DFDRNN consistently maintains excellent performance, demonstrating notable stability. This stability can be attributed to the mining capability of the SAM. Furthermore, the changes in average AUROC (Figure \ref{numk} (c)) and average AUPR (Figure \ref{numk} (d)) indicate that when $t = 7$, DFDRNN achieves the highest average AUROC of 0.9452 and the highest average AUPR of 0.5973 across the four datasets. For detailed numerical values, please refer to Supplementary Tables S1 and S2.

\subsection{Ablation analysis}

\subsubsection{Encoder}
The encoder of DFDRNN incorporates SAM to encode drugs (or diseases) into the similarity feature (SF) and the association feature (AF). We introduce two dual-feature extraction modules, SDDFE and CDDFE, to precisely encode drugs and diseases, with CDDFE changing the feature (CF) of drugs (or diseases). We developed various model variants to underscore these elements' importance in the DFDRNN encoder and evaluated their performance.

\begin{itemize}
    \item DFDRNN-SF: Remove the association feature and encode drugs and diseases solely as the similarity feature.
    \item DFDRNN-AF: Remove the similarity feature and encode drugs and diseases solely as the association feature.
    \item DFDRNN-GCN: Replace the SAM with a GCN to evaluate its effect on the model’s performance.
    \item DFDRNN-noCF: Remove the change of features in CDDFE, meaning that the similarity feature and the association feature will not be changed into each other after CDDFE.
\end{itemize}

Table \ref{Ablation analysis} presents the AUROC and AUPR values of DFDRNN and its four variants under 10-CV. By comparing the two variants of DFDRNN, DFDRNN-SF and DFDRNN-AF, we observe that DFDRNN-SF performs better than DFDRNN-AF on the Gdataset, Cdataset, and LRSSL, indicating the contribution of the similarity feature is greater than that of the association feature on these three datasets. However, on the Ldataset, the contribution of the association feature is greater than that of the similarity feature. Compared to DFDRNN, simultaneously encoding both features can significantly improve the model's performance. Compared to DFDRNN, DFDRNN-GCN demonstrates a noticeable decrease in AUROC values, highlighting the strong performance of SAM in extracting neighbor information. Furthermore, compared to DFDRNN, DFDRNN-noCF shows a significant decrease in AUPR on the Gdataset, Cdataset, and LRSSL datasets, and performs the worst on the Ldataset, indicating the necessity of changing features in the CDDFE module.

\begin{table*}[t]
\caption{The performance of DFDRNN and its four variants\label{Ablation analysis}}%
\tabcolsep=0pt%%
\centering
\begin{tabular*}{0.95\textwidth}{@{\extracolsep{\fill}}llccccccc@{\extracolsep{\fill}}}
\toprule%
\multirow{2}*{Variants} & Gdataset & & Cdataset & & Ldataset & & LRSSL & \\
\cline{2-3}\cline{4-5}\cline{6-7}\cline{8-9}\\
& AUROC & AUPR & AUROC & AUPR& AUROC & AUPR& AUROC & AUPR\\
\cline{1-9}\\
DFDRNN-SF  & 0.958 & 0.612 & \underline{0.972} & \textbf{0.690} & 0.873 & 0.538 & \underline{0.964} & 0.464\\
DFDRNN-AF  & 0.942 & 0.606 & 0.962 & 0.678 & \underline{0.884} & \underline{0.570} & 0.950 & \textbf{0.504}\\
DFDRNN-GCN  & 0.950 & \underline{0.620} & 0.964 & 0.671 & 0.874 & 0.551 & 0.956 & \underline{0.483}\\
DFDRNN-noCF & \underline{0.959} & 0.611 & 0.970 & 0.676 & 0.868 & 0.523 & 0.963 & 0.462\\
DFDRNN      & \textbf{0.960} & \textbf{0.622} & \textbf{0.973} & \underline{0.689} & \textbf{0.885} & \textbf{0.572} & \textbf{0.965} & \textbf{0.504}\\
\bottomrule
\end{tabular*}
\end{table*}

\subsubsection{Decoder}

\begin{figure}[h]%
\centering
\includegraphics[width=0.46\textwidth]{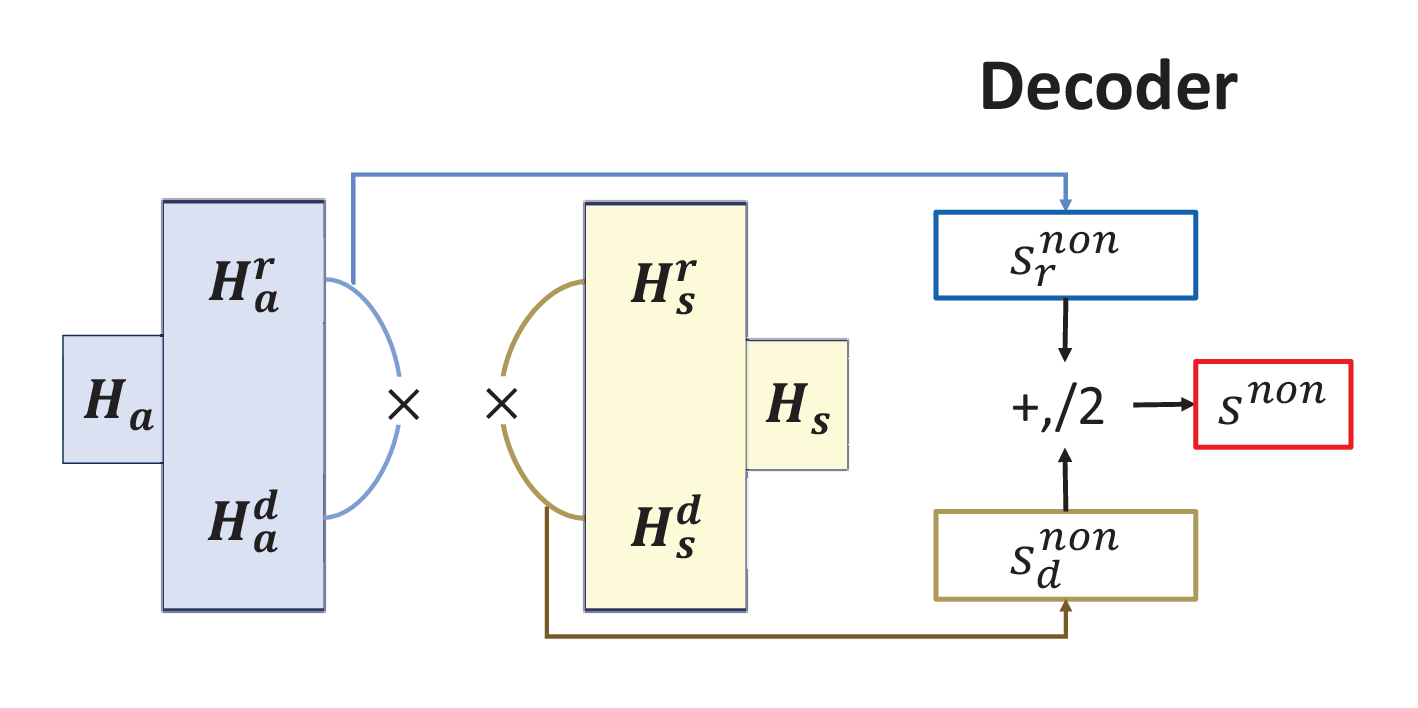}
\caption{non-cross dual-domain decoder}\label{non-cross}
\end{figure}

To demonstrate the effectiveness of the cross-dual-domain decoder, an experiment was designed to compare the AUROC value and the AUPR value obtained by DFDRNN with those obtained by five variants of DFDRNN, labeled as $s_r, s_d, s^{non}, s^{non}_r$, and $s^{non}_d$. Here, $s_r$ (Equation \ref{r}), $s_d$ (Equation \ref{d}), and $s^{non}$ represent the models derived from DFDRNN by substituting the cross-dual-domain decoder with a single drug-domain decoder, a single disease-domain decoder and a non-cross dual-domain decoder encompassing decoding results of both the drug and disease domains (as shown in Figure \ref{non-cross}), respectively. Similarly, $s^{non}_r$ and $s^{non}_d$ are models obtained from $s^{non}$ by replacing the non-cross dual-domain decoder with a single drug-domain decoder and a single disease-domain decoder, respectively. The specific implementation of the non-cross dual-domain decoder is described as follows.

\begin{alignat}{2}
 {s^{non}_r}&= \mathbf{sigmod}(  H^{r}_a\times (H^{d}_a)^T)\\
 {s^{non}_d}&= \mathbf{sigmod}(  H^{d}_s\times (H^{r}_s)^T)\\
 {s^{non}} &={(s^{non}_r + (^{non}_d)^T)/2.0}
\end{alignat}

\begin{figure*}[!t]
\begin{center}
\subfloat[]{
\includegraphics[width=0.46\textwidth]{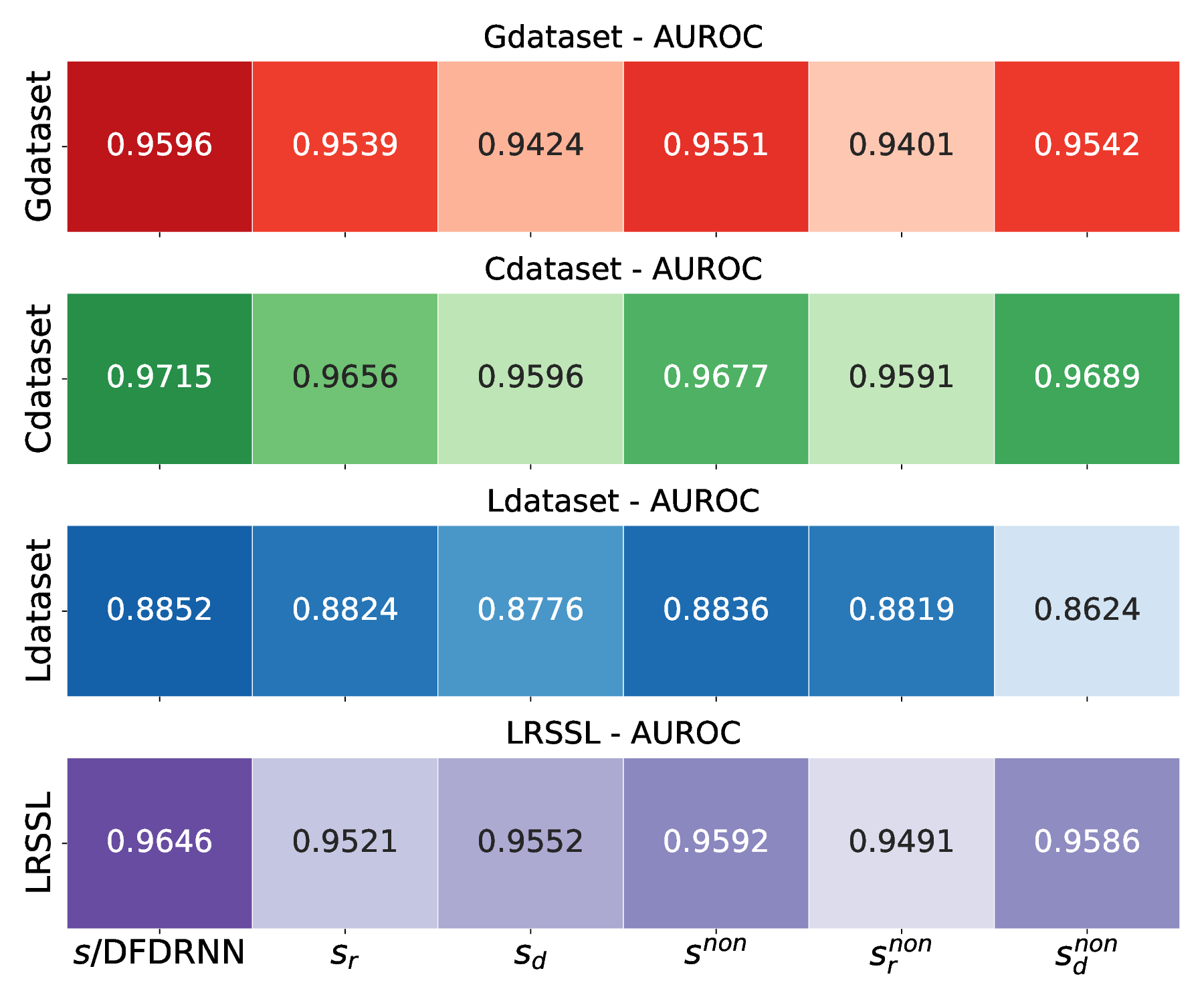}
}
\subfloat[]{
\includegraphics[width=0.46\textwidth]{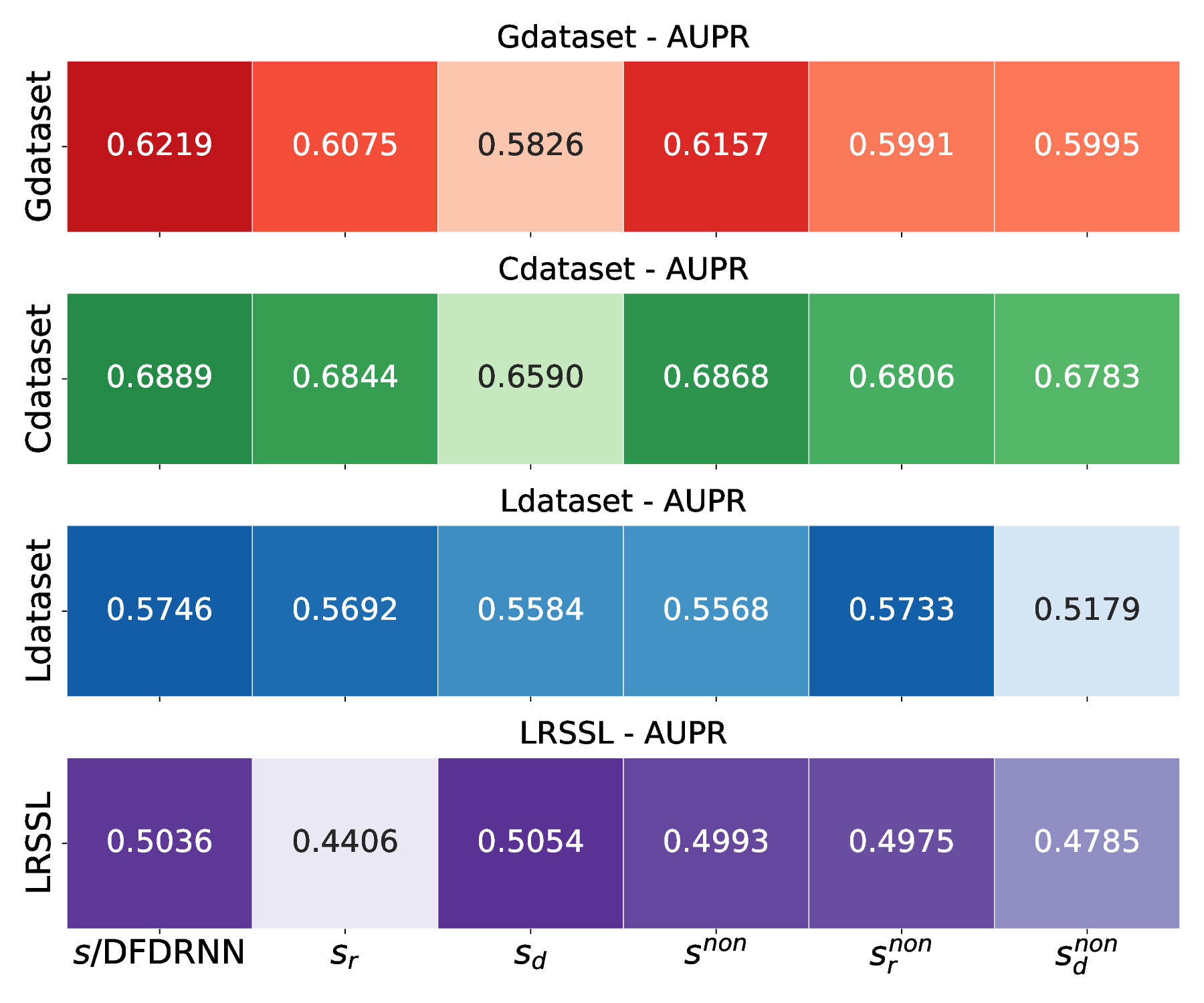}
}
\caption{Performance of different decoding methods: (a) AUROC heatmap; (b) AUPR heatmap.}\label{score}
\end{center}
\end{figure*}

The heatmaps, a crucial visual representation of the results, are shown in Figure \ref{score} across the four datasets: Gdataset, Cdataset, Ldataset, and LRSSL. DFDRNN's consistent outperformance in achieving the highest AUROC and AUPR values across all four datasets underscores the significance of the cross-dual-domain decoder. Furthermore, the performance of $ s_r$, $ s_d$, $ s^ {non}_r$, and $ s^ {non}_d$ consistently lags behind that of DFDRNN and $ s^ {non} $, further highlighting the effectiveness of dual-domain decoding.

\subsection{Identification of candidates for new diseases}
In late 2019, the discovery of COVID-19 presented a challenge in promptly researching and developing suitable drugs for treatment. Repurposing already approved drugs, such as remdesivir, favipiravir, and ribavirin, played a significant role in the timely control of the epidemic \cite{singh2020drug}. Finding suitable drugs for emerging diseases is a complex and crucial task. To assess DFDRNN's ability to predict candidate drugs for new diseases, we applied leave-one-out cross-validation (LOOCV) using the Gdataset. For each disease $d_x$, all drug-disease associations linked to $d_x$ were removed and treated as test data, while the remaining associations served as training data. The prediction relied on the similarity between $d_x$ and other diseases to identify its candidate drugs.
\begin{figure*}[!t]
\begin{center}
\subfloat[]{
\includegraphics[width=0.50\textwidth]{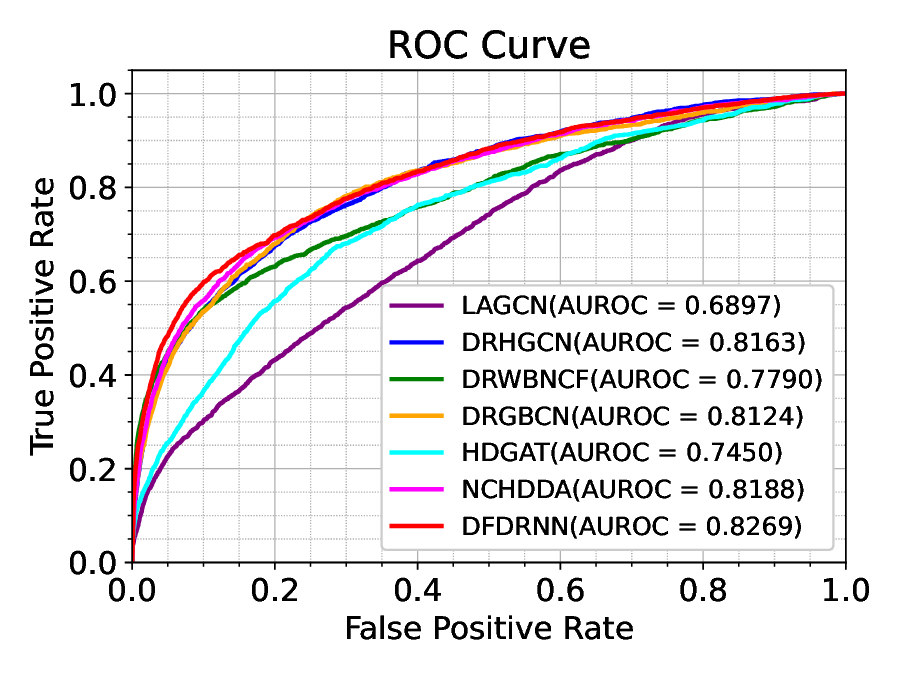}
}
\subfloat[]{
\includegraphics[width=0.50 \textwidth]{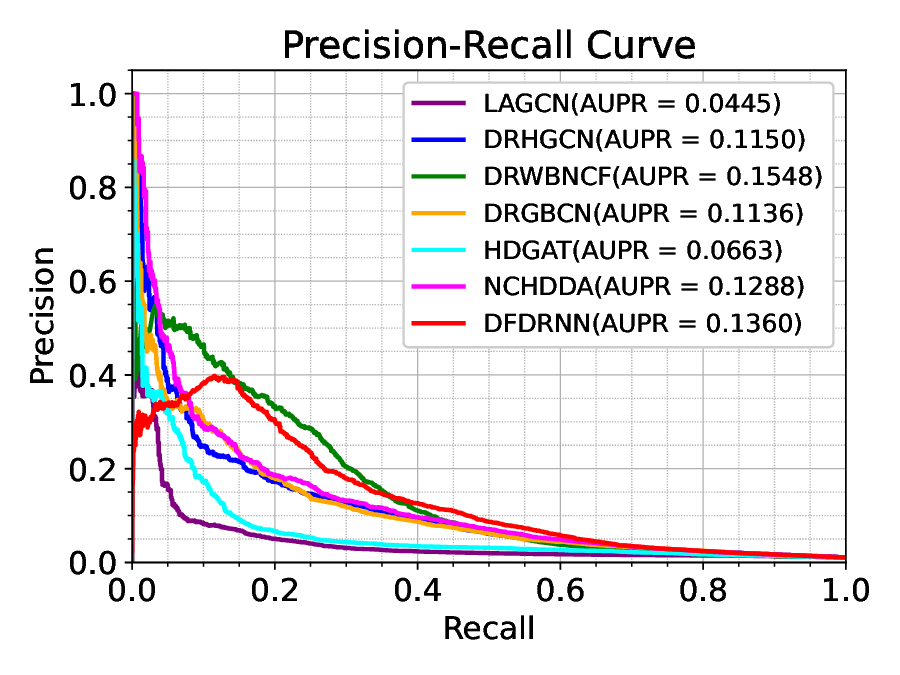}
}
\caption{The performance of all approaches in predicting candidate drugs for new diseases on Gdataset: (a) ROC curves illustrating the prediction outcomes of DFDRNN alongside other competitive models; (b) PR curves illustrating the prediction outcomes of DFDRNN alongside other competitive models.}\label{new_disease}
\end{center}
\end{figure*}

In Figure \ref{new_disease}, the performance of DFDRNN in identifying potential drugs for new diseases on the Gdataset is illustrated. The experimental results show that DFDRNN achieved the highest AUROC of 0.8269, surpassing the second model NCH-DDA by 0.81\%. Additionally, DFDRNN had the second-best performance in terms of AUPR, coming in behind only DRWBNCF.

\subsection{Case study}
To evaluate the practical effectiveness of DFDRNN, we performed case studies on two diseases, Alzheimer's disease (AD) and Parkinson's disease (PD), from the Gdataset. Using all drug-disease associations within the Gdataset as the training data, we trained DFDRNN to predict association probabilities for all unknown drug-disease pairs. We then ranked the drugs in descending order of predicted probabilities, identifying the top 10 candidate drugs for AD and PD. To validate DFDRNN's predictions, we referenced reliable sources and clinical trials, including DrugBank \cite{knox2024drugbank}, PubChem \cite{kim2016pubchem}, CTD \cite{davis2023CTD}, DrugCentral \cite{avram2021drugcentral}, and ClinicalTrials.

 AD, the most prevalent form of progressive dementia in the elderly, is a neurodegenerative disorder. Table \ref{AD} displays the top 10 drugs predicted by DFDRNN for AD. Among these, Amantadine is the first drug predicted to hold therapeutic potential for AD. Amantadine, commonly used as an antiviral and antiparkinson drug, exhibits actions as an antiviral, antiparkinson, dopaminergic, and analgesic agent. It can also be utilized for the prevention or symptomatic treatment of influenza A. Previous studies have shown improvements in the mental state of two AD patients following Amantadine treatment \cite{erkulwater1989amantadine}. Carbamazepine functions as an anticonvulsant and analgesic drug. Earlier studies have suggested that the combined therapy of carbamazepine and haloperidol holds promise in the clinical management of elderly AD patients \cite{lemke1995effect}. DFDRNN predicts an association with carbamazepine, which has been validated by CTD. Moreover, all the other drugs listed in Table \ref{AD} have been confirmed by relevant authoritative data sources, with a success rate of 100\%.

\begin{table}[ht]%
\centering
\caption{The top 10 candidate drugs for AD are ranked based on the DFDRNN prediction score}\label{AD}%
\begin{tabularx}{\textwidth}{@{\extracolsep{\fill}}lllX@{\extracolsep{\fill}}}
\toprule
Rank & Drug ID & Drug Name & Evidence\\
\midrule
1  & DB00915 & Amantadine     & DB/CTD/DrugCentral/ PubChem/ClinicalTrials \\
2  & DB00564 & Carbamazepine  & CTD \\
3  & DB04844 & Tetrabenazine  & CTD \\
4  & DB00682 & Warfarin       & CTD \\
5  & DB01241 & Gemfibrozil    & CTD/PubChem/ ClinicalTrials \\
6  & DB00160 & Alanine        & CTD/ClinicalTrials \\
7  & DB00903 & Etacrynic acid & CTD \\
8  & DB00928 & Azacitidine    & CTD \\
9  & DB00482 & Celecoxib      & CTD \\
10 & DB01262 & Decitabine     & CTD \\
\bottomrule
\end{tabularx}
\end{table}

\begin{table}[ht]%
\centering
\caption{The top 10 candidate drugs for PD are ranked based on the DFDRNN prediction score}\label{PD}%
\begin{tabularx}{\textwidth}{@{\extracolsep{\fill}}lllX@{\extracolsep{\fill}}}
\toprule
Rank & Drug ID & Drug Name & Evidence\\
\midrule
1  & DB00502 & Haloperidol     & DB/CTD/PubChem/ ClinicalTrials \\
2  & DB01068 & Clonazepam      & CTD/ClinicalTrials \\
3  & DB01235 & Levodopa        & DB/CTD/DrugCentral/ PubChem/ClinicalTrials\\
4  & DB00810 & Biperiden       & DB/CTD/DrugCentral/ PubChem \\
5  & DB00424 & Hyoscyamine     & DB \\
6  & DB00575 & Clonidine       & ClinicalTrials \\
7  & DB00915 & Amantadine      & DB/DrugCentral/PubCh-em/ClinicalTrials \\
8  & DB00989 & Rivastigmine    & DB/CTD/DrugCentral/ PubChem/ClinicalTrials \\
9  & DB00376 & Trihexyphenidyl & DB/CTD/DrugCentral/ PubChem \\
10 & DB00215 & Citalopram      & CTD \\
\bottomrule
\end{tabularx}
\end{table}

Parkinson's disease (PD) is the second most common neurogenic disorder after AD. Table \ref {PD}  presents the top 10 potential candidate drugs for PD predicted by DFDRNN. Among these candidates, Haloperidol, an antipsychotic agent, has been confirmed to be associated with PD by DB, CTD, PubChem, and ClinicalTrials data sources. Additionally, Levodopa, a dopamine precursor used in PD management, has also been confirmed by multiple data sources. The fact that the predictions made by DFDRNN for the top 10 PD-related drugs have been validated by authoritative data sources provides reassurance about the reliability of the model.

\section{Conclusion}\label{section4}
Our study introduces DFDRNN, a neural network model designed for drug repositioning. DFDRNN uses the SDDFE and CDDFE modules to extract the similarity and association features for drugs (or diseases), respectively, ensuring more appropriate encoding of drugs (or diseases). Compared with the SDDFE module, the CDDFE module changes the similarity (or association) feature into the association (or similarity) feature. In both modules, we use the SAM to mining the neighbor information of drugs (and diseases). Additionally, the DFDRNN model fuses three similarity (or association) features into a new similarity (or association) feature and uses a layer attention mechanism to calculate the final similarity (or association) feature. Finally, a cross-dual-domain decoder process is adopted to predict drug-disease associations in both the drug and disease domains. DFDRNN performs best in 10-CV across four benchmark datasets and achieves the highest AUROC in LOOCV. The experimental results confirm the effectiveness of our proposed DFDRNN in drug repositioning. 

For future work, we would like to investigate whether specific enhancements, such as integrating temporal data and handling multi-modal data, could further enhance the performance of DFDRNN. By incorporating temporal data, the model can better understand and predict patterns over time. Integration of data from different modalities could provide a more comprehensive understanding of the input. We aim to identify and implement the appropriate techniques and models for these enhancements. Furthermore, we plan to investigate incorporating other biological entities, such as miRNAs and proteins, for drug repositioning.

\begin{acknowledgement}

The authors thank the National Key R\&D Program of China (No. 2021ZD0112400) and the National Natural Science Foundation of China (Grants 62172072 and 62272115) for their financial support.

\end{acknowledgement}

\section{Data Availability}

The implementation and data of the DFDRNN model are available at \url{https://github.com/siekersun/DFDRNN.git}.

\begin{suppinfo}

\begin{itemize}
  \item manuscript.pdf: Manuscript PDF File
  \item manuscript.zip: This zip package contains files for an academic paper written in LaTeX, including the main document, references, and document class files for formatting.
  \item supplementary.docx:Supporting Information for Publication
\end{itemize}

\end{suppinfo}

\bibliography{main}

\end{document}